\documentclass[pdflatex,sn-apa]{sn-jnl}
\usepackage[utf8]{inputenc}
\usepackage{graphicx}
\usepackage{booktabs}
\usepackage{multirow}
\usepackage{array}
\usepackage{tabularray}
\usepackage{subcaption}
\usepackage{amsmath,amssymb,amsfonts}
\usepackage{amsthm}
\usepackage{mathrsfs}
\usepackage[title]{appendix}
\usepackage{xcolor}
\usepackage{textcomp}
\usepackage{siunitx}
\usepackage{listings}
\usepackage{algorithm}
\usepackage{algpseudocode}
\usepackage{natbib}
\usepackage{csquotes}
\usepackage{placeins}
\usepackage{float}
\usepackage{xurl}
\usepackage{url}
\makeatletter
\@ifundefined{APACrefURL}{}{%
  \renewcommand{\APACrefURL}[1]{\url{#1}}
}
\makeatother

\begin{document}

\title[Article Title]{A Novel Approach to Tomato Harvesting Using a Hybrid Gripper with Semantic Segmentation and Keypoint Detection}


\author*[1]{\fnm{Shahid} \sur{Ansari}}\email{ansari.shahid.b2@tohoku.ac.jp}

\author[2]{\fnm{Mahendra Kumar} \sur{Gohil}}\email{mahendraku2014@gmail.com}

\author[3]{\fnm{Yusuke} \sur{Maeda}}\email{maeda@ynu.ac.jp}

\author[4]{\fnm{Bishakh} \sur{Bhattacharya}}\email{bishakh@iitk.ac.in}

\affil[1]{\orgdiv{Department of Mechanical and Aerospace Engineering}, \orgname{Tohoku University},
\orgaddress{\city{Sendai}, \state{Miyagi}, \postcode{980-8579}, \country{Japan}}}

\affil[2]{\orgname{TSTS},
\orgaddress{\city{Kanpur}, \state{Uttar Pradesh}, \postcode{208016}, \country{India}}}

\affil[3]{\orgdiv{Division of Systems Research, Faculty of Engineering}, \orgname{Yokohama National University},
\orgaddress{\city{Yokohama}, \state{Kanagawa}, \postcode{240-0067}, \country{Japan}}}

\affil[4]{\orgdiv{Department of Mechanical Engineering}, \orgname{Indian Institute of Technology Kanpur (IIT Kanpur)},
\orgaddress{\city{Kanpur}, \state{Uttar Pradesh}, \postcode{208016}, \country{India}}}

\abstract{
This paper presents an autonomous tomato-harvesting system built around a hybrid robotic gripper that combines six soft auxetic fingers with a rigid exoskeleton and a latex basket to achieve gentle, cage-like grasping. The gripper is driven by a servo-actuated Scotch--yoke mechanism, and includes separator leaves that form a conical frustum for fruit isolation, with an integrated micro-servo cutter for pedicel cutting. For perception, an RGB--D camera and a Detectron2-based pipeline perform semantic segmentation of ripe/unripe tomatoes and keypoint localization of the pedicel and fruit center under occlusion and variable illumination. An analytical model derived using the principle of virtual work relates servo torque to grasp force, enabling design-level reasoning about actuation requirements. During execution, closed-loop grasp-force regulation is achieved using a proportional--integral--derivative controller with feedback from force-sensitive resistors mounted on selected fingers to prevent slip and bruising. Motion execution is supported by Particle Swarm Optimization (PSO)--based trajectory planning for a 5-DOF manipulator. Experiments demonstrate complete picking cycles (approach, separation, cutting, grasping, transport, release) with an average cycle time of 24.34~s and an overall success rate of approximately 80\%, while maintaining low grasp forces (0.20--0.50~N). These results validate the proposed hybrid gripper and integrated vision--control pipeline for reliable harvesting in cluttered environments.
}

\maketitle

\section{Introduction}\label{sec1}

Precision agriculture and smart farming are increasingly adopted to improve productivity, reduce input waste, and maintain high product quality under growing demand. These approaches integrate sensing, automation, and data-driven decision-making to improve crop yield and post-harvest quality (\cite{gupta2020security}). In this context, autonomous robotic harvesting is a key enabling technology for horticulture, where labor shortages and high labor costs directly affect production and consistency.

Despite progress in mechanization, many conventional harvesting methods (e.g., combine harvesters, reapers, and trunk shakers) are unsuitable for soft and delicate crops such as tomatoes and strawberries because large contact forces and impacts can bruise or damage the fruit (\cite{shojaei2021intelligent,cho2014using}). Selective harvesting, where fruits are picked individually at the appropriate ripeness stage, is therefore preferred for high-value crops. However, selective harvesting remains challenging because a robot must (i) detect the target fruit under occlusion, (ii) estimate its pose and identify the pedicel cutting location, and (iii) execute grasping and detachment without damaging the fruit or plant. In real cultivation environments, tomatoes are often densely packed and partially occluded by leaves and branches, making perception and reliable manipulation difficult (\cite{chen2015reasoning}). Consequently, integrated harvesting systems that combine compliant end-effectors, robust perception, and closed-loop control remain an active research topic (\cite{comba2010robotics,ling2019dual}).

A wide range of end-effectors has been explored for harvesting and handling soft produce. (\cite{gao2022development}) proposed a clamping-type end effector actuated through a pneumatic mechanism to generate finger rotation.Soft and hybrid grippers have been studied to improve conformal contact and reduce pressure hotspots. \cite{tawk20223d} presented a 3D-printed modular soft pneumatic gripper based on mechanical metamaterials, while \cite{Kaur2019toward} described a metamaterial-inspired robotic finger with compliant auxetic joints and embedded sensing. Recent work has also investigated hybrid meta-grippers for tomato harvesting and analyzed how auxetic lattice orientation affects grasp conformability and force/torque requirements (\cite{ansari2025hybrid})
. For tomato harvesting, \cite{kondo2010development} developed an end-effector capable of harvesting individual tomatoes and clusters and sensing the peduncle using strain sensors. \cite{ansari2022design} proposed a cage-like soft gripper with a separation mechanism and a servo-driven iris-based pedicel-cutting unit. \cite{liu2018soft} introduced an underactuated, sensor-less soft gripper module integrated with machine vision for fruit picking. More generally, anthropomorphic and tendon-driven approaches have also been proposed for fragile-object handling; for example, \cite{baker2023star} presented a soft robotic gripper driven by twisted string actuators.

Alongside end-effector design, perception is a major bottleneck in cluttered crops. Semantic segmentation and keypoint detection have become important computer vision techniques for fruit detection, instance delineation, and picking-point localization. \cite{Zhou2023} used YOLOv7 for dragon fruit localization and a PSP-Ellipse method to identify endpoints. \cite{Liang2022} combined semantic segmentation (BiSeNet V2) with a pruned YOLOv4 network for real-time assessment of defective apples. \cite{10.1109/ICRA46639.2022.9812303} introduced strawberry datasets with picking-point annotations and demonstrated keypoint detection for grasping and harvesting tasks. \cite{Yan2023} proposed keypoint-based grasping and cutting-point estimation within an instance-segmentation framework for pumpkin harvesting.

Motivated by these challenges, this work presents a complete tomato-harvesting robotic system that integrates a soft--rigid hybrid gripper with deep learning--based perception and motion/control for end-to-end harvesting. The main objectives of this research are:
\begin{enumerate}
    \item To design and analyze a hybrid gripper that combines six soft auxetic fingers with a rigid exoskeleton for gentle yet firm grasping, and to derive a torque--force relationship using the principle of virtual work for actuation sizing.
    \item To evaluate mechanical and control performance across the essential harvesting stages: approach, separation, cutting, grasping, transport, and release.
    \item To develop a vision-based perception module employing deep learning--based semantic segmentation to distinguish ripe and unripe tomatoes and keypoint detection to localize the tomato center and pedicel under occlusion and variable illumination.
    \item To implement PSO-based trajectory planning for a 5-DOF robotic arm and closed-loop PID force control of the gripper for safe and adaptive tomato handling.
\end{enumerate}
\section{Design and analysis}\label{sec:2}
\begin{figure}[!htbp]
    \centering
    \includegraphics[width=\linewidth]{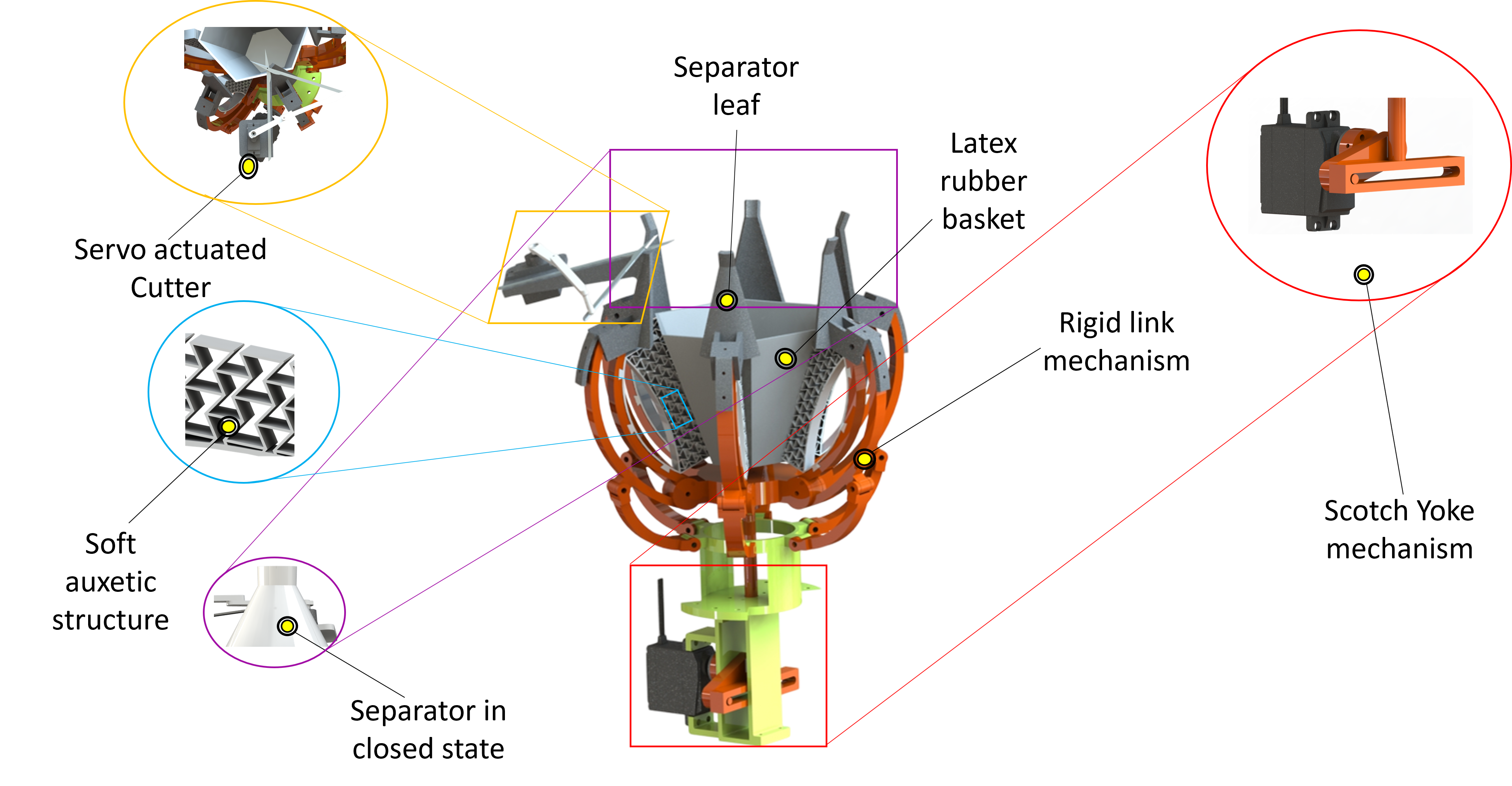}
    \caption{CAD model of the proposed hybrid gripper. Left: cutter mechanism actuated by a micro-servo (top), soft auxetic finger structures for gentle grasping (middle), and separator leaves forming a conical frustum during separation (bottom). Right: servo-actuated Scotch--yoke mechanism used to drive finger closure.}
    \label{fig:gripperassembly_CAD}
\end{figure}

The proposed gripper combines a 2D re-entrant auxetic structure (metamaterial) with rigid linkages that form an internal exoskeleton, as shown in Fig.~\ref{fig:gripperassembly_CAD}. 
Auxetic structures exhibit a negative Poisson's ratio and can expand laterally when stretched due to their re-entrant geometry, which enables compliant, conformal contact compared with conventional lattices \cite{mir2014review}. This behavior is beneficial for grasping soft and deformable produce because it can reduce localized stress concentrations and distribute contact pressure more uniformly over the fruit surface.

\begin{figure}[!htbp]
    \centering
    \includegraphics[width=0.8\linewidth]{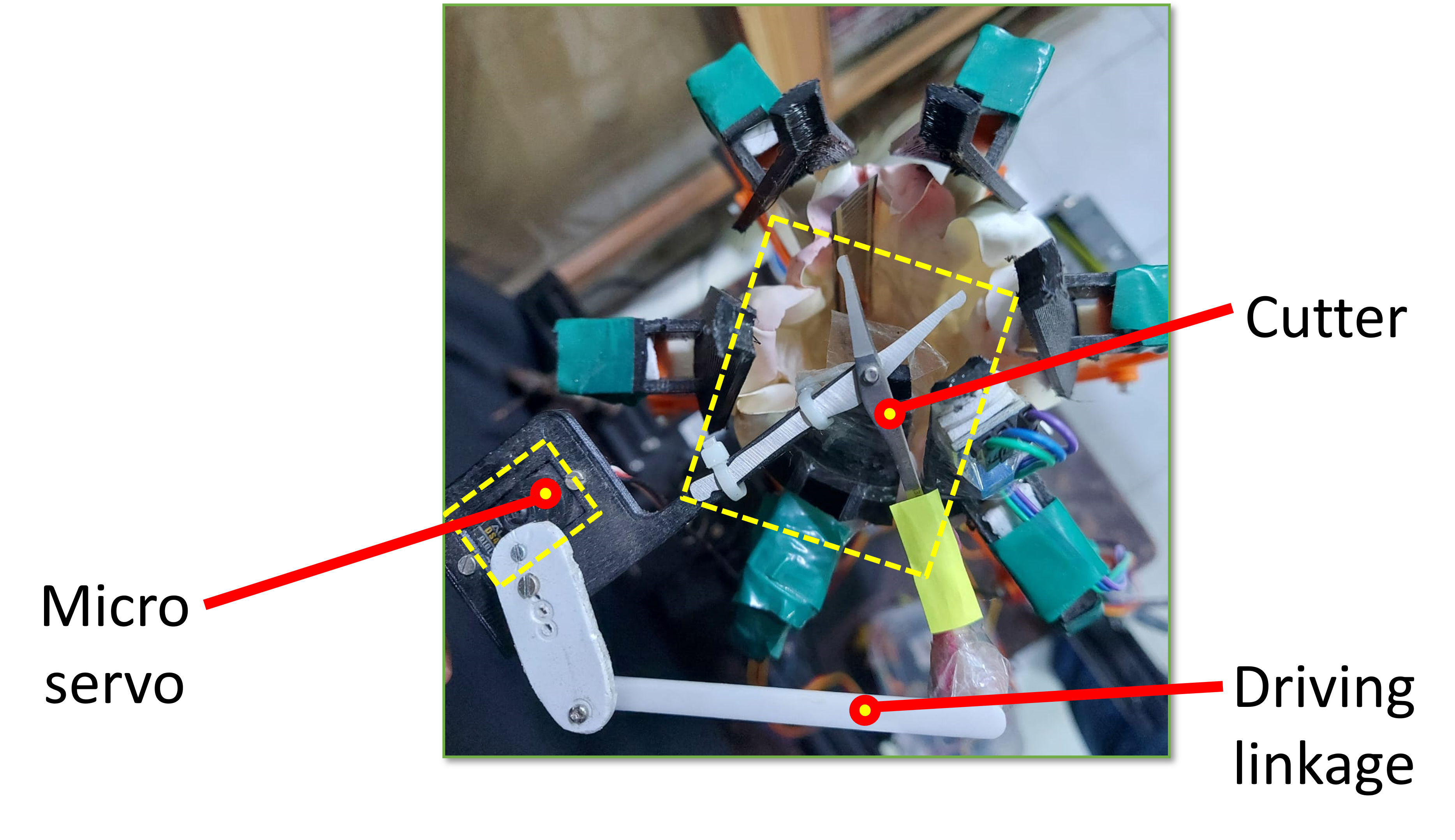}
    \caption{Micro-servo-actuated cutter mechanism mounted on a separator leaf. The cutter is actuated when the separator leaves are fully closed and the pedicel is aligned within the cutting zone.}
    \label{fig:cutter_mechanism}
\end{figure}

The design objective is to achieve a stable hold using a gentle caging grasp produced by six symmetric fingers. Each finger integrates a re-entrant honeycomb pattern with a curved, leaf-spring-like element to provide both shape adaptivity and effective stiffness. To prevent fruit escape during transport, a thin latex basket is bonded over the auxetic structures, forming a compliant capture volume around the tomato. In clustered fruit scenarios, six separator leaves close into a conical frustum and are used to isolate the target tomato by guiding neighboring tomatoes outward. After separation, a micro-servo-actuated cutter mounted on one separator leaf cuts the pedicel, as shown in Fig.~\ref{fig:cutter_mechanism}.
\subsection{Materials and components}\label{sec:materials}
The outer rigid links and the Scotch--yoke drive mechanism are 3D printed using PLA (polylactic acid) on a Creality Ender-3 V2 printer, while the compliant auxetic structures are 3D printed using TPU (thermoplastic polyurethane) on an Ultimaker system. The holding basket is fabricated from \SI{1}{mm} latex rubber, providing mild elastic resistance during opening while helping retain the fruit during transport. The gripper drive motor is an Orange OT5316M \SI{7.4}{V} metal-gear digital servo motor, and the cutter is actuated using an Align DS426M digital micro-servo. For perception, a ZED2i RGB--D camera (Stereolabs) is used for tomato detection and keypoint localization. Manipulation is performed using a 5-DOF ViperX-300 robotic arm (Interbotix).
\subsection{Torque requirement calculation}\label{sec:torque_req}
\begin{figure}[!htbp]
    \centering
    \includegraphics[width=0.75\linewidth]{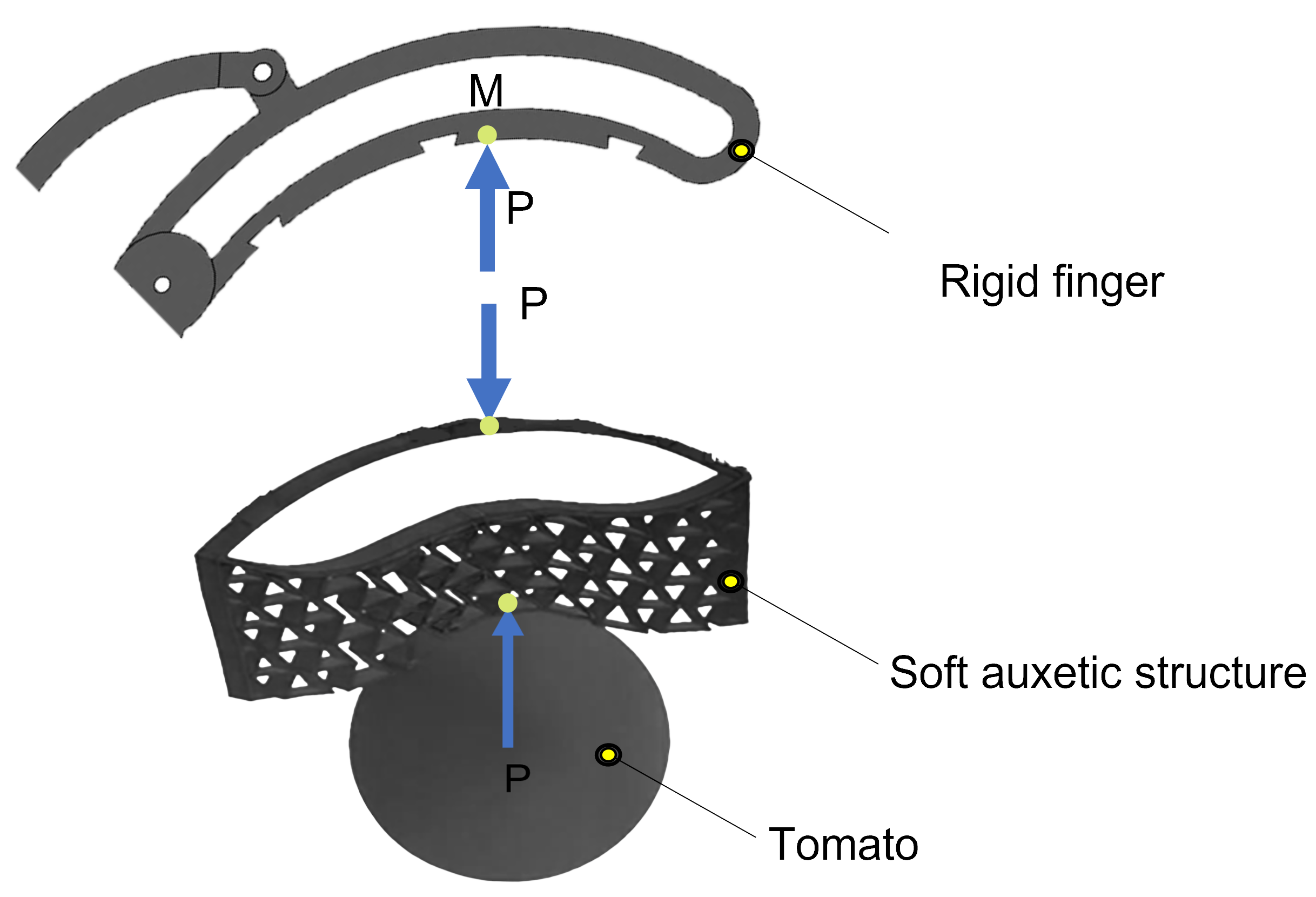}
    \caption{Representative finger contact model used for static force and virtual-work-based torque analysis.}
    \label{fig:static_force}
\end{figure}

\begin{figure}[!htbp]
    \centering
    \includegraphics[width=0.7\linewidth]{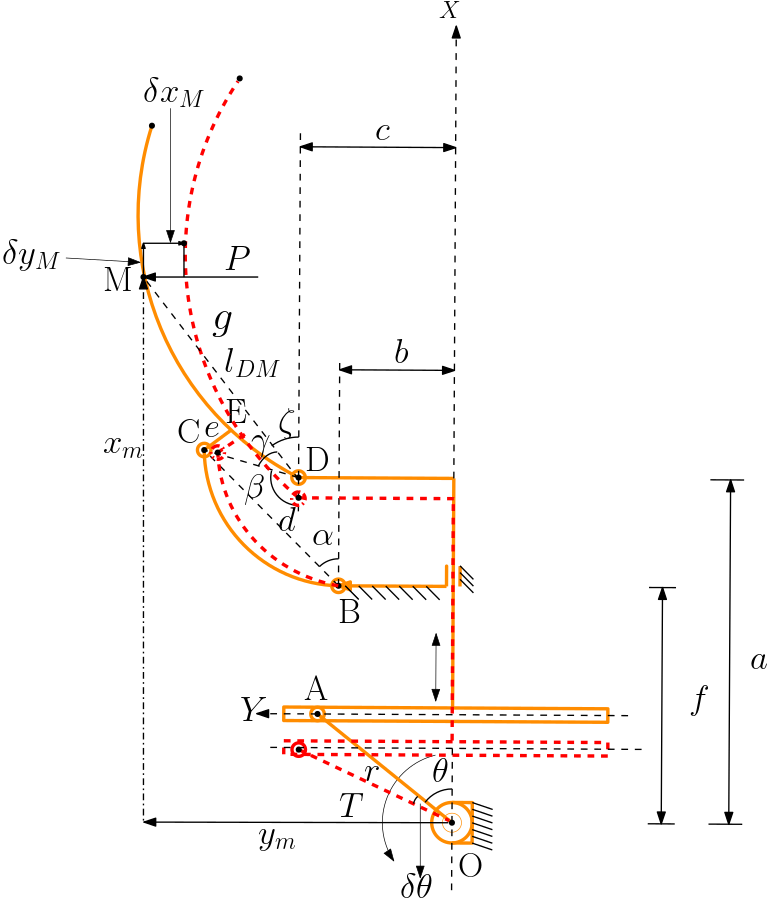}
    \caption{Kinematic and geometric parameters used in the torque analysis. The reaction force arises through the rigid linkage supported by the auxetic structure during contact.}
    \label{fig:gripper_kinematics}
\end{figure}

To estimate the actuator torque required to grasp a tomato using the auxetic-structure-based fingers (Fig.~\ref{fig:static_force}),we use the principle of virtual work to derive an input--output relationship.
 For an ideal mechanism in static equilibrium, the total virtual work is zero for all admissible virtual displacements:
\begin{equation}
\delta U = 0 .
\label{eq:virtual_work}
\end{equation}
The kinematic parameters used in the analysis are shown in Fig.~\ref{fig:gripper_kinematics}.
A representative virtual-work balance can be written as:
\begin{gather}
T\,\delta\theta + P\,\delta x_m + P\,\delta y_m = 0, \label{eq:vw_balance}\\
x_m = r\cos\theta + l_s + l_{DM}\cos\xi, \qquad 
y_m = l_p + l_{DM}\sin\xi, \label{eq:vw_pos}\\
\delta x_m = -r\sin\theta\,\delta\theta - l_{DM}\sin\xi\,\delta\xi, \qquad
\delta y_m = l_{DM}\cos\xi\,\delta\xi, \label{eq:vw_delta}\\
r\cos\theta + f = a + e\cos\beta + d\cos\xi, \qquad
c + e\sin\beta = b + d\sin\xi, \label{eq:vw_loop}\\
\xi = 180^{\circ} - \beta - \gamma, \qquad
k = \sqrt{(r\cos\theta + f - a)^2 + (c - b)^2}, \label{eq:vw_k}\\
u = \tan^{-1}\!\left(\frac{c - b}{r\cos\theta + f - a}\right), \label{eq:vw_u}\\
\beta = \cos^{-1}\!\left(\frac{(r\cos\theta + f - a)^2 + (c - b)^2 - d^2}{2ek}\right) - u, \label{eq:vw_beta}\\
T = P\,l_{DM}\sin\xi\,\frac{\delta\xi}{\delta\theta} + P\,r\sin\theta. \label{eq:torque_relation}
\end{gather}

\begin{figure}[!htbp]
    \centering
    \includegraphics[width=\linewidth]{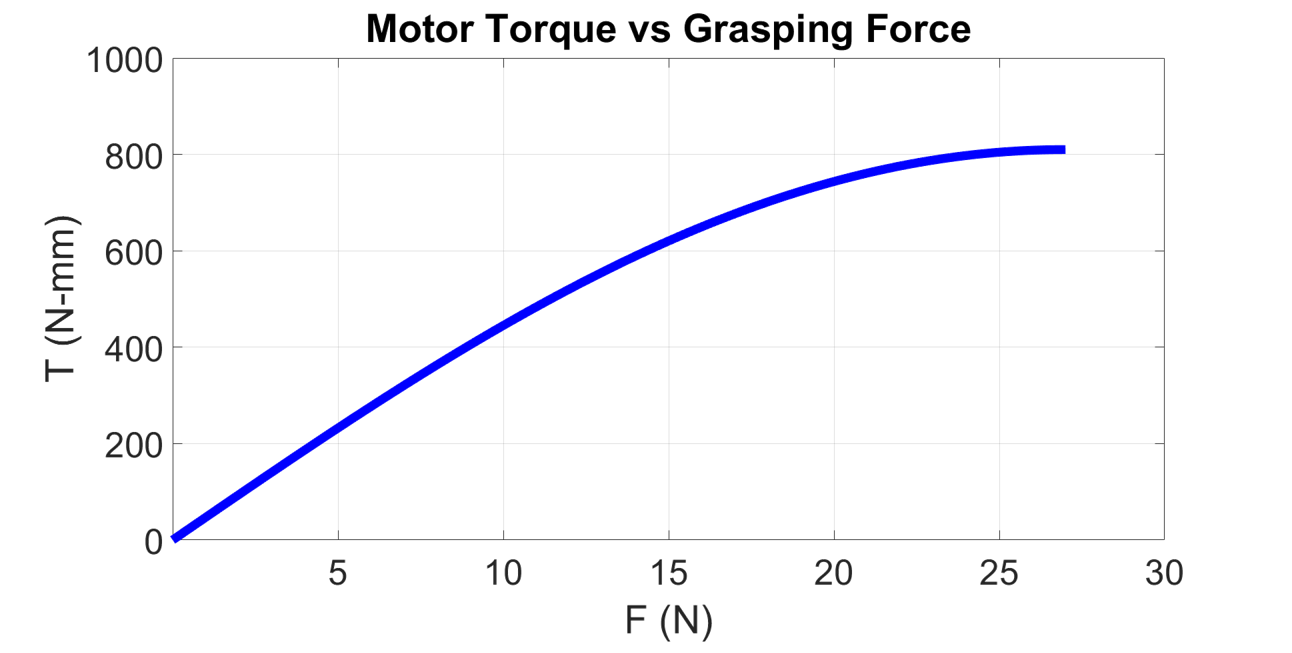}
    \caption{Nonlinear variation of motor torque with the desired grasping force for the gripper fingers.}
    \label{fig:Force_torque_curve}
\end{figure}

Here, $T$ is the motor torque (N$\cdot$mm), $P$ is the transmitted force (N), $x_m$ and $y_m$ are the driven point coordinates, and $r$, $a$, $b$, $c$, $d$, $e$, $f$, $k$, $l_p$, and $l_{DM}$ are geometric parameters in millimeters, while $\beta$, $\gamma$, $\theta$, $\xi$, and $u$ are angles in degrees. Equations~\eqref{eq:vw_balance}--\eqref{eq:torque_relation} provide the relationship between the motor torque and the effective transmitted force for a representative finger.The complete derivation of the virtual-work-based torque--force relationship used in this section is provided in appendix~\ref{app:vw_derivation}.
 For symmetric closure of the six-finger gripper, total actuation demand increases relative to a single finger due to simultaneous contact and transmission losses as can be depicted from Fig. \ref{fig:Force_torque_curve}. Therefore, actuator selection should account for multi-finger loading and mechanical inefficiencies rather than assuming ideal linear scaling.

\section{Sensing and control}\label{sec:3}

The electronic hardware of the hybrid auxetic gripper integrates sensing, embedded control, and actuation to enable adaptive tomato harvesting. Three force-sensitive resistor (FSR) strips are placed on selected auxetic finger surfaces to measure grasp contact force. Each FSR is interfaced via a voltage-divider circuit and read by an ATmega328P microcontroller. Although the end-effector has six fingers, only three fingers are instrumented to reduce wiring and computational complexity; because the mechanism closes approximately symmetrically under nominal alignment, the measured forces provide a practical estimate of grasp-force level for feedback control. Sensor data are acquired using PLX-DAQ for monitoring and controller validation, and are used to reduce slip risk while limiting excessive contact forces that may bruise the fruit. An infrared sensor assists in detecting the tomato position near the cutting region, while a ZED~2i RGB--D camera provides depth feedback for guiding the manipulator. The control architecture includes a master controller and two ATmega328P-based slave units for sensor acquisition and actuation interfacing. Separate voltage regulation circuits supply the servos that drive the cutter and the gripper mechanism to maintain stable operation under varying load. Together, these subsystems enable coordinated sensing, control, and actuation for reliable tomato-harvesting experiments \cite{ansari2025design}.

\begin{figure}[!htbp]
    \centering
    \includegraphics[width=0.8\linewidth]{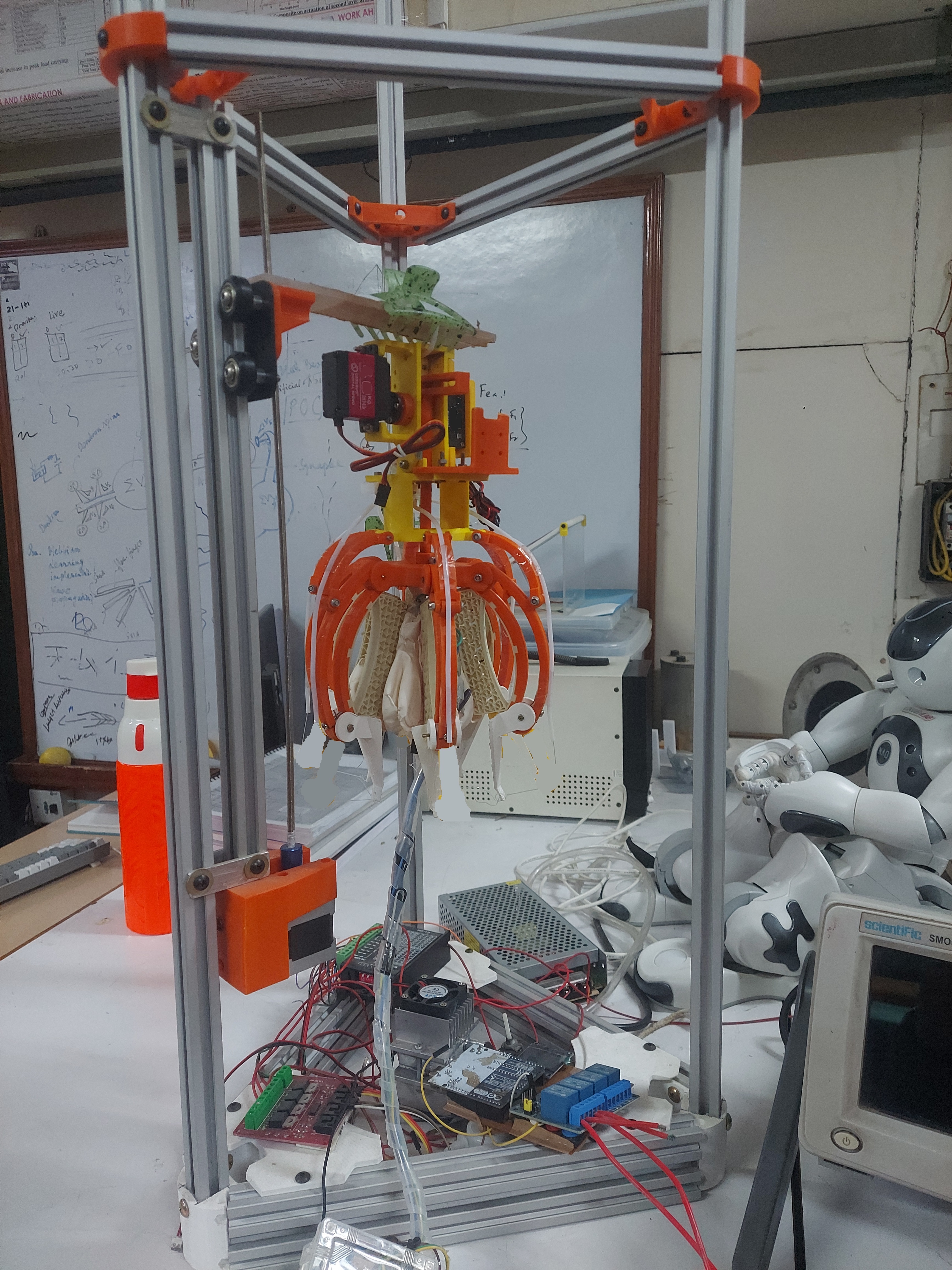}
    \caption{Tabletop experimental setup used for grasp-force regulation experiments and characterization.}
    \label{fig:Grasping-tester-tabletop}
\end{figure}

The grasping experiments were conducted using a tabletop setup (Fig.~\ref{fig:Grasping-tester-tabletop}) to ensure repeatability and to allow controlled evaluation of the force-regulation behavior under varying tomato sizes.

\subsection{PID gain selection and tuning for the hybrid auxetic gripper}\label{sec:pid_gain_selection}

\begin{figure}[!htbp]
    \centering
    \includegraphics[width=\linewidth]{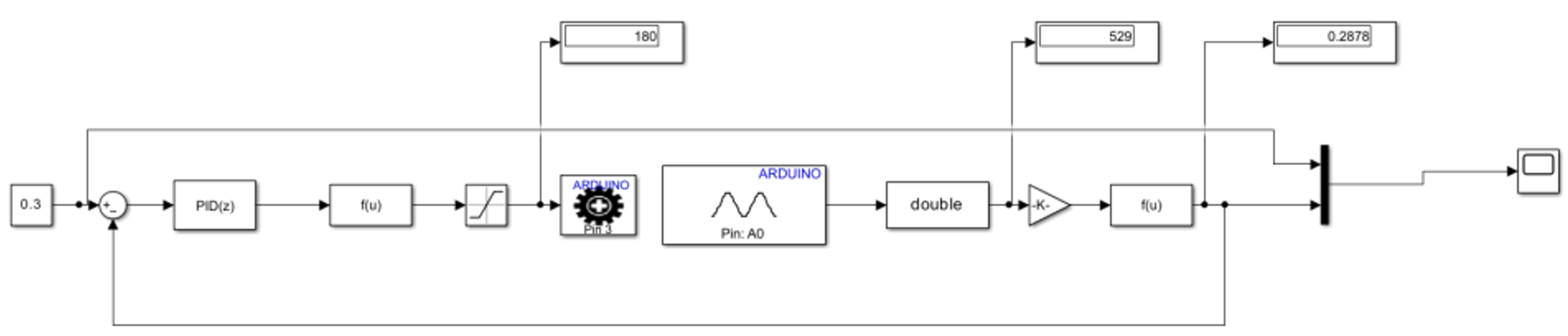}
    \caption{Simulink controller block diagram for grasp-force regulation. The measured FSR force is compared to the reference force; the discrete PID output is mapped and saturated to a safe servo-angle range for the Scotch--yoke-driven gripper.}
    \label{fig:simulink_block_controller}
\end{figure}

For delicate fruit handling, the controller must apply sufficient grip force to prevent slip while avoiding excessive pressure that can bruise the tomato. A closed-loop proportional--integral--derivative (PID) controller is used to regulate the grasping force of the hybrid auxetic gripper using real-time FSR feedback, as illustrated in Fig.~\ref{fig:simulink_block_controller}. The control objective is to track a desired reference force while maintaining stable contact during the hold phase.

PID control is selected because it is computationally efficient for real-time implementation and provides an effective balance between responsiveness and stability. The proportional term reduces instantaneous force error, the integral term compensates steady-state offsets caused by friction or sensor bias, and the derivative term damps transient force spikes during first contact. This is particularly relevant for the proposed gripper because the compliant auxetic structures introduce nonlinear contact dynamics and variability across fruit size and surface conditions.

\subsection{Initial gain estimates and tuning strategy}\label{sec:pid_tuning}
PID gains were determined experimentally using a two-stage procedure. First, a Ziegler--Nichols-style initialization was applied by setting $K_i$ and $K_d$ to zero and gradually increasing $K_p$ until sustained oscillations appeared around the target force. The resulting ultimate gain and oscillation period were then used to obtain an initial set of gains.

Next, the gains were refined manually by incrementally increasing $K_i$ to reduce steady-state error and introducing a small $K_d$ term to limit overshoot and damp oscillations during initial contact. This refinement is important because the FSR signals are noise-prone and the gripper--tomato interaction is inherently nonlinear.

\subsection{Empirical gain selection}\label{sec:pid_gains}
The gains that provided stable force tracking with low overshoot in the presented setup were:
\begin{equation}
\begin{aligned}
K_p &= 0.15 \;(^\circ/\mathrm{N}),\\
K_i &= 0.02 \;(^\circ/(\mathrm{N}\cdot\mathrm{s})),\\
K_d &= 0.001 \;(^\circ\cdot\mathrm{s}/\mathrm{N}).
\end{aligned}
\end{equation}
These gains were validated through repeated trials using a hobby-grade servo rated for 5--6~V \cite{miuzei_ds3218_datasheet}, three FSR sensors (approximately up to 2~N range), and re-entrant honeycomb auxetic finger structures. The selected $K_p$ provides a sufficiently fast response without inducing sustained oscillations, $K_i$ compensates steady-state offsets while limiting integrator windup, and the small $K_d$ term suppresses transient force spikes without amplifying sensor noise.

To regulate grasping force during harvesting, the controller receives a reference force $F_{\mathrm{ref}}$ (set to \SI{0.30}{N} in experiments) and compares it with the measured force $F(t)$ from the FSR sensors to compute the error $e(t)=F_{\mathrm{ref}}-F(t)$. A discrete PID controller generates the control output, which is transformed (scaled and saturated) into a servo-angle command (typically 0--180$^\circ$) and sent to the gripper servo through Arduino PWM. The FSR signal is read through an analog input (e.g., A0), converted from ADC counts to force using an experimentally calibrated mapping, and fed back to close the loop.

\subsection{Performance analysis}\label{sec:pid_performance}
\begin{figure}[!htbp]
    \centering
    \includegraphics[width=\linewidth]{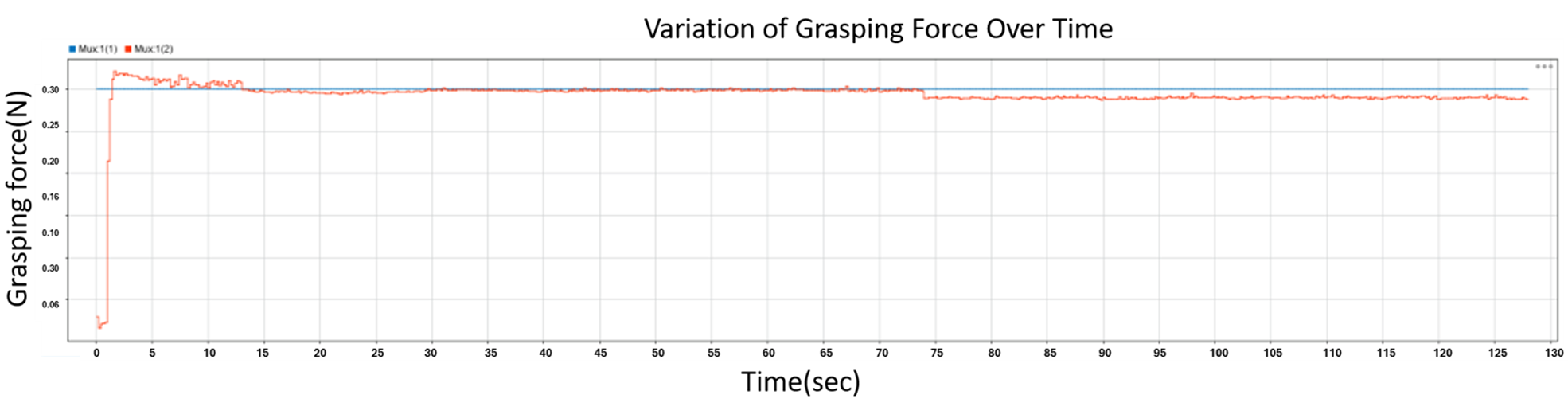}
    \caption{PID force-regulation response. The reference force is shown in blue and the measured grasp force is shown in red.}
    \label{fig:simulink_controller_response}
\end{figure}

With these gains, the force response in Fig.~\ref{fig:simulink_controller_response} shows that the grasping force converges to the reference value of \SI{0.30}{N} within approximately \SIrange{1}{2}{s}, with peak overshoot remaining below about \SI{10}{\percent} of $F_{\mathrm{ref}}$. During the hold phase, the controller maintains stable regulation, keeping the steady-state deviation within approximately \SI{0.02}{N} across repeated trials.

The reported gains are validated for the specific servo, transmission, sensor characteristics, and auxetic geometry used in this study. For different actuators or transmission ratios, $K_p$ may need re-scaling; for different sensor noise levels, $K_d$ and filtering may need adjustment; and changes in auxetic geometry or material can alter effective compliance and require re-tuning. Overall, the chosen gains provide a practical baseline for force-controlled gripping using auxetic finger structures in delicate produce handling.
\section{Force variation and gripping performance analysis}\label{sec:force_variation}
\begin{figure}[!htbp]
    \centering
    \includegraphics[width=\linewidth]{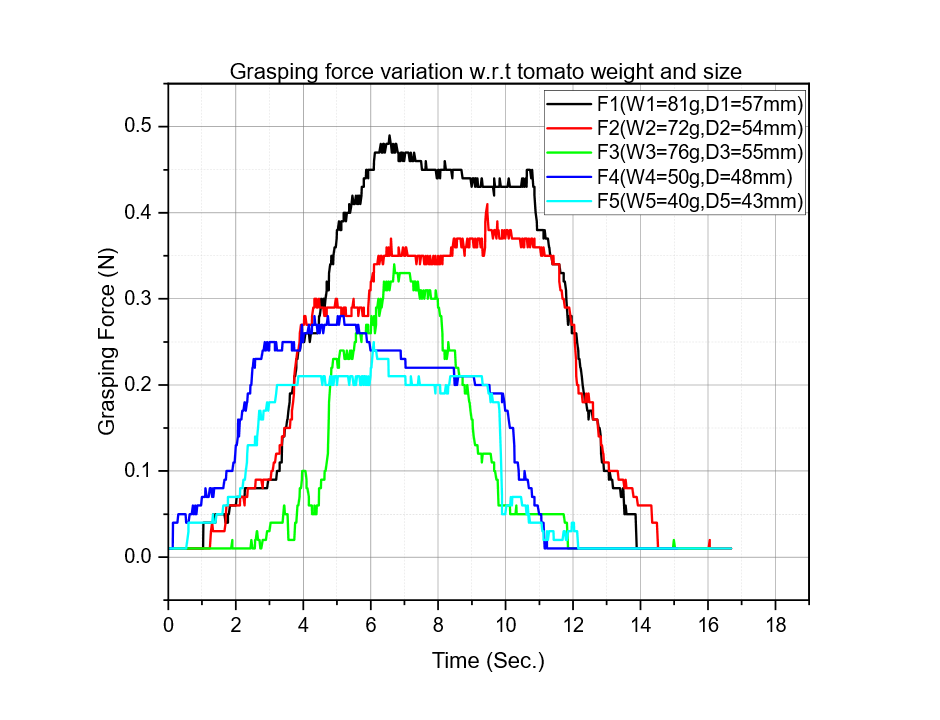}
    \caption{Grasp-force versus time measured using three FSR strips during grasp--hold--release trials.
The plotted force is the average of the three calibrated FSR signals at each time instant.
Traces $F_1$--$F_5$ correspond to five tomato samples with different weights and diameters.}
    \label{fig:tomato_forces}
\end{figure}

The representative force--time profiles measured during grasp trials are shown in Fig.~\ref{fig:tomato_forces}. At each time step, the plotted grasping force is the average of the three instrumented FSR strips,
i.e., $F(t)=\frac{1}{3}\sum_{i=1}^{3}F_i(t)$, where $F_i(t)$ denotes the calibrated contact force from sensor $i$.
Because the six-finger mechanism closes approximately symmetrically, this averaged measurement provides a practical estimate of the global grasp-force level used for feedback control. 
Five tomato samples with different weights (40--81~g) and diameters (43--57~mm) were evaluated. The trials are denoted as $F_1$ (81~g, 57~mm), $F_2$ (72~g, 54~mm), $F_3$ (76~g, 55~mm), $F_4$ (50~g, 48~mm), and $F_5$ (40~g, 43~mm). Each profile exhibits a consistent grasp--hold--release pattern, while peak magnitudes and rise rates vary with tomato size and mass.
Heavier/larger tomatoes (e.g., $F_1$) require higher steady grasp forces, reaching approximately 0.48--0.50~N around 5--7~s, whereas smaller/lighter tomatoes (e.g., $F_5$) stabilize at approximately 0.20--0.25~N. In all cases, the force increases rapidly during the initial contact phase (0--2~s) as the rigid linkage closes and the auxetic structure begins to conform to the fruit surface. A quasi-steady plateau is observed during the hold interval (approximately 5--10~s), followed by a smooth decay during controlled release. The absence of abrupt force jumps during the hold and release phases indicates stable force regulation and repeatable actuation behavior under the tested conditions.

These profiles reflect the role of the re-entrant auxetic geometry and the curved leaf-spring-like compliance within each finger. As the mechanism closes, the auxetic segments deform to match local curvature, increasing the effective contact area and helping reduce localized pressure concentrations. In addition, the six-finger caging configuration provides multi-point contact, improving grasp robustness at relatively low force levels. While the present study reports force signals (rather than direct pressure/bruise metrics), maintaining force within the 0.20--0.50~N range across tested tomato sizes supports the goal of gentle yet stable handling for selective harvesting.

From a design standpoint, the results suggest that the same end-effector can accommodate tomatoes within the tested size range with minimal change in control settings: larger fruits naturally lead to slightly higher equilibrium forces due to increased deformation and contact engagement, while smaller fruits are held at lower force levels.

\subsection{Tomato semantic segmentation with tomato-center and pedicel keypoint detection}\label{sec:seg_kpt}
\begin{figure}[!htbp]
    \centering
    \includegraphics[width=0.75\columnwidth]{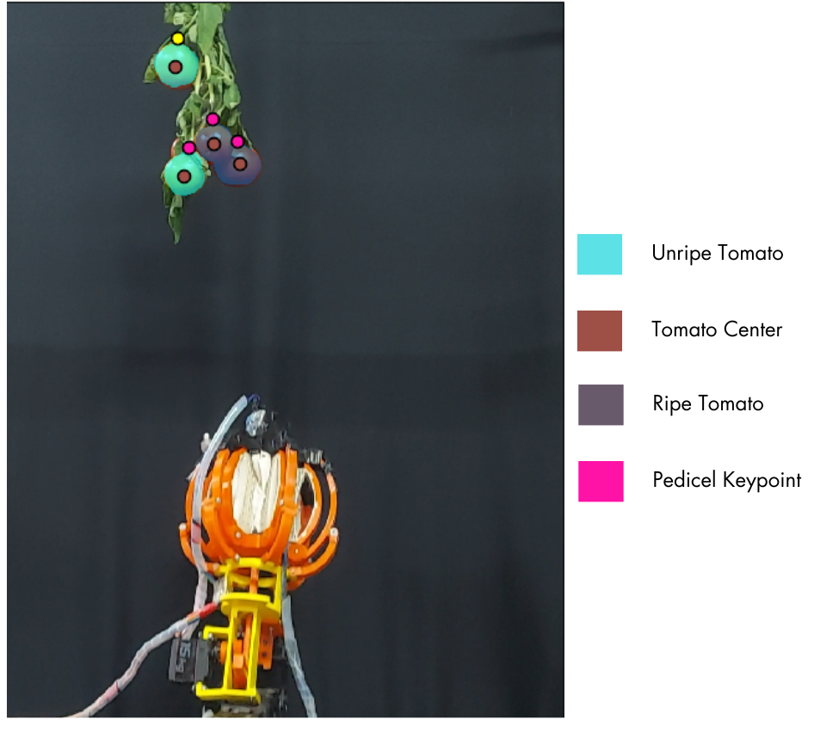}
    \caption{Example outputs of the perception module: semantic segmentation of tomato instances (ripe/unripe) and predicted keypoints for the pedicel and tomato center.}
    \label{fig:tomato_detection}
\end{figure}

To enable autonomous harvesting, a perception module is developed for (i) semantic segmentation of ripe and unripe tomatoes and (ii) keypoint localization of the tomato center and pedicel. Tomato-center localization supports target pose estimation and approach planning, while pedicel keypoint localization supports cutter alignment and cutting-point estimation. The training data include the Rob2Pheno dataset \cite{afonso21} along with additional field and laboratory images collected in this study. Detectron2 \cite{wu2019detectron2} is used to train models for segmentation and keypoint detection. As shown in Fig.~\ref{fig:tomato_detection}, the trained model outputs instance masks for ripe/unripe tomatoes and predicts keypoints corresponding to the pedicel and tomato center, which are subsequently used for cutter alignment and approach planning.

\noindent\textbf{Evaluation metrics.}
For the perception module, we report COCO-style average precision (AP) computed over IoU thresholds from 0.50 to 0.95 with a step of 0.05, denoted as AP@[0.50:0.95]. In Table~\ref{tab:detectron2_results}, \emph{mAP} refers to the mean \textbf{instance segmentation mask AP} (Mask AP@[0.50:0.95]) averaged across the tomato classes. Precision and recall are computed on the test set using the matched predicted instances under the same IoU-based assignment protocol.

Models are trained using the Adam optimizer with an initial learning rate of 0.001 for 20{,}000 iterations. Multiple backbone networks are evaluated, and the comparative results are summarized in Table~\ref{tab:detectron2_results}. In our experiments, the ResNeXt-based backbone achieved the strongest overall performance across the reported metrics.

\begin{table*}[!htbp]
\caption{Performance comparison of Detectron2 configurations with different backbone networks.}
\label{tab:detectron2_results}
\centering
\resizebox{\linewidth}{!}{%
\begin{tabular}{lcccccccc}
\toprule
Backbone & Iterations & Learning rate & Optimizer & Train loss & Test loss & mAP & Precision & Recall \\
\midrule
ResNet-50  & 20{,}000 & 0.001 & Adam & 0.20 & 0.30 & 0.75 & 0.80 & 0.70 \\
ResNeXt    & 20{,}000 & 0.001 & Adam & 0.10 & 0.20 & 0.80 & 0.85 & 0.75 \\
ResNet-101 & 20{,}000 & 0.001 & Adam & 0.15 & 0.25 & 0.77 & 0.82 & 0.72 \\
\bottomrule
\end{tabular}%
}
\end{table*}

\section{Experimental evaluation}\label{sec:experiments}

\begin{figure}[!htbp]
    \centering
    \includegraphics[width=\linewidth]{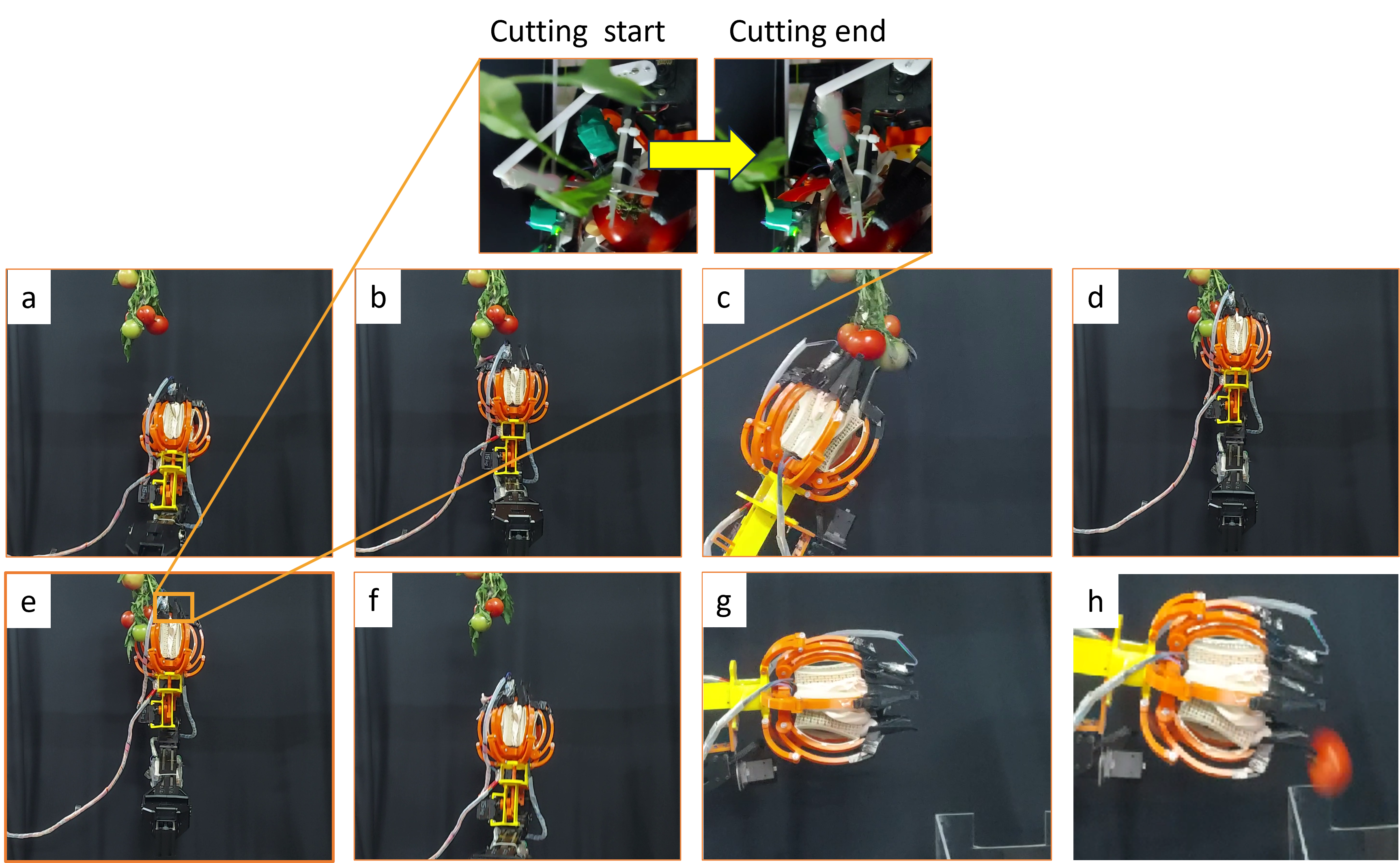}
    \caption{Experimental sequence of a complete picking cycle in laboratory conditions:
    (a--b) approach toward the tomato cluster,
    (c--d) separation of the target tomato using the conical-frustum separator leaves,
    (e) pedicel cutting using the micro-servo-driven cutter,
    (f) gentle caging grasp with the auxetic fingers and latex basket,
    (g) departure toward the punnet,
    (h) release into the punnet.}
    \label{fig:picking_operation}
\end{figure}

A complete tomato-picking cycle was evaluated in a controlled laboratory setting. The perception module first processes the RGB--D stream and returns (i) the tomato instance segmentation and (ii) two keypoints: the tomato center (used as the approach target) and the pedicel (used to align the cutter). Using these outputs, the robot arm executes the following sequence:

\begin{enumerate}
    \item \textbf{Approach:} The end-effector is guided toward the target tomato using the estimated 3D position derived from RGB--D data.
    \item \textbf{Separation:} The separator leaves are actuated to form a conical frustum that gently pushes aside neighboring tomatoes and foliage, allowing the target tomato to enter the gripper workspace.
    \item \textbf{Cutting:} Once the tomato reaches a predefined depth in the gripper, the distance between the predicted pedicel keypoint and the cutter reference position is evaluated. When this distance meets the cutting criterion, the micro-servo actuates the cutter to sever the pedicel.
    \item \textbf{Grasping and transport:} After cutting, the tomato is retained inside the flexible latex basket and enclosed by the six auxetic fingers. Closed-loop force control is applied during the grasp/hold phase to maintain gentle contact.
    \item \textbf{Placement:} The manipulator moves to the punnet and releases the tomato in a controlled manner.
\end{enumerate}
Figure~\ref{fig:picking_operation} illustrates the main stages of the implemented harvesting procedure. The cutting image is captured separately (not simultaneously) to provide clear visualization of pedicel severing, since a multi-camera setup was not used to record the process from multiple viewpoints.

\begin{figure}[!htbp]
    \centering
    \includegraphics[width=0.9\linewidth]{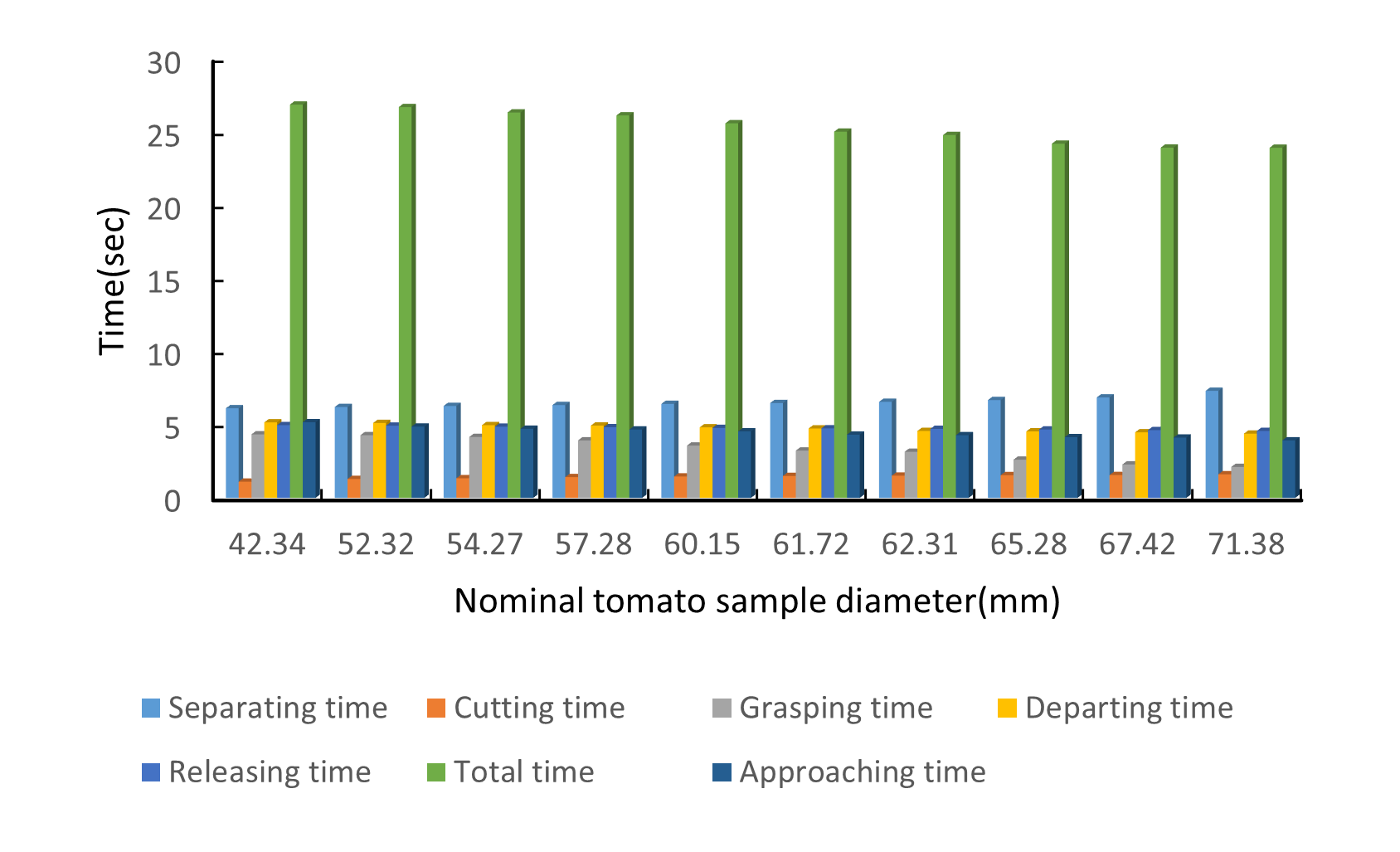}
    \caption{Time breakdown of the picking cycle across tomato samples of different diameters. The bars indicate the duration of approach, separation, cutting, grasping, departure, release, and the total cycle time.}
    \label{fig:picking_time_breakdown}
\end{figure}

Figure~\ref{fig:picking_time_breakdown} summarizes the time required for each stage of the picking cycle across the tested tomato diameters. The outcome of each trial and the corresponding failure modes are summarized in Table~\ref{tab:trial_success}.

Across ten experimental trials, the system achieved an overall picking success rate of approximately 80\% under the reported laboratory conditions.\begin{table*}[!htbp]
\centering
\caption{Outcome summary of ten tomato-picking trials. A trial is marked \emph{Success} if the system completes separation, pedicel cutting, grasping/transport, and placement into the punnet without dropping the fruit.}
\label{tab:trial_success}
\begin{tabular}{c c p{0.62\linewidth}}
\toprule
Trial \# & Outcome & Failure reason (if any) \\
\midrule
1  & Success & -- \\
2  & Success & -- \\
3  & Fail    & Pedicel misalignment; the cutter did not fully sever the pedicel \\
4  & Success & -- \\
5  & Success & -- \\
6  & Fail    & Occlusion-induced keypoint error caused cutter actuation at an incorrect depth \\
7  & Success & -- \\
8  & Success & -- \\
9  & Fail    & Tomato slipped during transfer due to insufficient contact force at the onset of motion \\
10 & Success & -- \\
\bottomrule
\end{tabular}
\end{table*}

Overall, 8 out of 10 trials were successful, corresponding to an average picking success rate of 80\%. A trial was considered successful if the system completed separation, pedicel cutting, secure grasping/transport, and placement into the punnet without dropping the fruit.

\section{Conclusion}\label{sec:conclusion}
This paper presented an integrated robotic system for tomato harvesting that combines a hybrid soft--rigid end-effector, deep-learning-based perception, and motion execution on a robotic manipulator. The proposed gripper employs six auxetic soft fingers supported by a rigid exoskeleton to realize a gentle caging grasp suitable for delicate produce. A mechanical analysis based on the principle of virtual work was developed to relate actuator torque to grasping force, providing design-level insight into actuation requirements. For perception, a Detectron2-based pipeline performs ripe/unripe tomato segmentation and keypoint localization of the pedicel and tomato center to support target selection, approach planning, and cutter alignment. During manipulation, closed-loop grasp-force regulation using FSR feedback enables stable holding at low force levels, while PSO-based trajectory planning is used to generate feasible arm motions for the harvesting sequence.

Experimental evaluation in laboratory conditions demonstrated complete picking cycles (approach, separation, pedicel cutting, grasping, transport, and release) with an average cycle time of approximately 24.3~s and an overall success rate of about 80\% over ten trials.

The measured grasping forces remained within 0.20--0.50~N across the tested tomato sizes, indicating that the end-effector can maintain secure grasps while limiting contact forces. These results support the feasibility of the proposed hybrid gripper and integrated perception--control framework for selective harvesting in cluttered scenes.

Future work will focus on (i) improving the robustness of pedicel cutting and alignment under occlusion, (ii) enhancing perception reliability under outdoor illumination and background variability, (iii) expanding evaluation to broader fruit varieties and larger-scale field trials with quantitative damage/bruise assessment, and (iv) accelerating cycle time through improved planning and execution strategies for multi-fruit harvesting.

\bmhead{Acknowledgements}
This research was supported by the Department of Science and Technology (DST), Government of India, under Project No.~DST/ME/2020009.

\section*{Author Contributions}
Shahid Ansari (SA): Conceptualization, methodology, investigation, experimental design and execution, data collection, writing---original draft, and manuscript revision. 
Mahendra Kumar Gohil (MG): Control methodology, controller design and implementation, experimental support, data interpretation, and manuscript revision. 
Bishakh Bhattacharya (BB): Supervision, resources, and writing---review and editing. 
Yusuke Maeda (YM): Writing---review and editing.

\section*{Declarations}
\subsection*{Competing interests}
The authors declare that they have no known competing financial interests or personal relationships that could have appeared to influence the work reported in this paper.

\bibstyle{sn-apa}
\bibliography{sn-bibliography}
\appendix \label{app:vw_derivation}
\section{Mathematical Derivation for the relation between the required torque to provide the desired grasping force}
Using the principle of virtual work, the gripper is analyzed under an infinitesimal configuration change driven by the motor torque, while the grasping reaction force $P$ acts on the rigid finger link through the auxetic structure.
\begin{gather*}
    T\delta\theta+P\delta x_{m}+P\delta y_{m}=0 \tag{1}\\
x_{m}=r\cos\theta+l_{s}+l_{DM}\cos\xi \tag{2}\\
y_{m}=l_{p}+l_{DM}\sin \xi \tag{3}\\
\delta x_{m} = -r\sin \theta \delta \theta-l_{DM}\sin \xi\\
\delta y_{m} = l_{DM}\cos \xi \delta \xi\\
T\delta \theta+P(-r\sin \theta \delta \theta-l_{DM}\sin\xi \delta \xi)-Pl_{DM}\cos\xi\delta\xi=0\\
(T-Pr\sin\theta)\delta\theta-Pl_{DM}(\cos\xi-\sin\xi)\delta\xi=0\\
T=Pl_{DM}(\cos\xi-\sin\xi)\frac{\delta\xi}{\delta\theta}+Pr\sin\theta\\
rcos\theta+f=a+e\cos\beta+d\cos\xi \tag{4}\\
c+e\sin\beta=b+d\sin\xi \tag{5}\\
\beta+\gamma+\delta=180^o\\
\xi=180^o-\beta-\gamma \tag{6}\\
d\cos\xi=(r\cos\theta+f-a-e\cos\beta)\\
d\sin\xi=(c+e\sin\beta-b)\\
d^2=\underbrace{(c+esin\beta-b)^2}_{\text{Part:1}}+\underbrace{(r\cos\theta+f-a-e\cos\beta)^2}_{\text{Part:2}}\\
Part1:
(c-b)^2+(e\sin\beta)^2+2(c-b)e\sin\beta\\
Part2:(r\cos\theta+f-a)^2+(e\sin\beta)^2-2(r\cos\theta+f-a)e\cos\beta\\
Part1+Part2:\\
(r\cos\theta+f-a)^2+(c-b)^2+e^2+\\\underbrace{2e[\underbrace{-(r\cos\theta+f-a)}_{\text{$k\cos u$}}\cos\beta+\underbrace{(c-b)}_{\text{$k\sin u$}}\sin\beta]}_{\text{$-2ek\cos(\beta-u)$}}\\
(r\cos\theta+f-a)=k\cos u\\
(c-b)=k\sin u\\
k=\sqrt{(r\cos\theta+f-a)^2+(c-b)^2} \tag{7}\\
\cos(\beta+u)=\frac{1}{2ek}[(r\cos\theta+f-a)^2+(c-b)^2-d^2]\\
\beta=\cos^{-1}\frac{1}{2ek}[(r\cos\theta+f-a)^2+(c-b)^2-d^2]-u \tag{8}\\
\tan u=\frac{c-b}{r\cos\theta+f-a}\\
u=\tan^{-1}(\frac{c-b}{r\cos\theta+f-a}) \tag{9}\\
\boxed{T =Pl_{DM}\sin\xi\frac{\delta\xi}{\delta\theta}+Pr\sin\theta}
\end{gather*}
\end{document}